# Measuring Outcomes in Healthcare Economics using Artificial Intelligence: with Application to Resource Management

Chih-Hao Huang[1]
E-mail: CHUANG21@GMU.EDU
ORCiD: 0000-0003-1655-8482

Feras A. Batarseh*[2] *(corresponding author)*
E-mail: BATARSEH@VT.EDU
ORCiD: 0000-0002-6062-2747

Adel Boueiz[3]
E-mail: AELBOUEIZ@MGH.HARVARD.EDU
ORCiD: 0000-0003-1638-8575

Ajay Kulkarni[1]
E-mail: AKULKAR8@GMU.EDU
ORCiD: 0000-0002-3620-2670

Po-Hsuan Su[1]
E-mail: PSU4@GMU.EDU

Jahan Aman[1]
E-mail: JAMAN@GMU.EDU

*College of Science, George Mason University (GMU), Fairfax, VA 22030 USA[1]*

*Bradley Department of Electrical and Computer Engineering, Virginia Polytechnic Institute and State University (Virginia Tech), Arlington, VA 22203 USA[2]*

*Channing Division of Network Medicine; Pulmonary and Critical Care Medicine, Brigham and Women's Hospital, Harvard Medical School, Boston, MA 02115 USA[3]*

**Abstract:** The quality of service in healthcare is constantly challenged by outlier events such as pandemics (i.e. Covid-19) and natural disasters (such as hurricanes and earthquakes). In most cases, such events lead to critical uncertainties in decision making, as well as in multiple medical and economic aspects at a hospital. External (geographic) or internal factors (medical and managerial), lead to shifts in planning and budgeting, but most importantly, reduces confidence in conventional processes. In some cases, support from other hospitals proves necessary, which exacerbates the planning aspect. This manuscript presents three data-driven methods that provide data-driven indicators to help healthcare managers organize their economics and identify the most optimum plan for resources allocation and sharing. Conventional decision-making methods fall short in recommending validated policies for managers. Using reinforcement learning, genetic algorithms, traveling salesman, and clustering, we experimented with different healthcare variables and presented tools and outcomes that could be applied at health institutes. Experiments are performed; the results are recorded, evaluated, and presented.

**Policy Significance Statement:** Conventional decision-making methods fall short in recommending validated and evidence-based policies during outlier events. During such events, Artificial Intelligence (AI) methods provide a viable alternative for better decision and policy making at medical institutions. We present three methods that provide data-driven guidance during healthcare crises; the methods aim at helping hospital managers and public health officials in preparing their budgets, identifying optimized economic plans for distribution of health relevant items (such as vaccines and masks), and overall resource allocation.

## 1. Introduction and Motivation

Healthcare resources consist of materials, personnel, facilities, and anything else that can be used for providing a healthcare service to patients at hospitals. Studies in healthcare economics have suffered from lack of relevance to practical realities on the ground. Potential deployments of such economic re-evaluations have proven that the reality is different from theory [1]. In the wake of a global pandemic, policy making has been scrutinized and public trust in opinion-based policy making has been diminished. It is more important than ever to begin integrating AI systems into the policy evaluation and advocacy process. Evidence-based approaches facilitate policy making in a manner that is more relative to the scientific method, beginning with initial hypotheses and evaluating them based on experimentation. Accordingly, in this study, we compare: (1) conventional (unnormalized, no learning, no optimization), with (2) AI-based (normalized, unsupervised learning, with optimization) methods, and derive recommendations to hospital managers accordingly. Healthcare management and executive teams often face the challenge of resources' management but end up compromising decisions to avoid issues with public health standards/laws and other safety compliances. That strategy is however destined to fail due to the many hidden consequences; for example, a survey identified that most finance directors in Scottish hospitals were unaware of the cost of back injuries among nurses [2]. The vast majority of hospitals run their budgets and economics based on previous experiences and heuristic intuitions of management. Moreover, the Economic Rate of Return (ERR) is very difficult to quantify in the context of health, given the multi-dimensional aspects that feed into the budget. For instance, patient satisfaction, readmission rates, hospital costs and budgets, and many other factors are important to consider during the measurement of the efficiency and quality of service.

Moreover, decision making's difficulty exacerbates during outlier events that directly affect healthcare services. Multiple factors influencing healthcare resource allocation in such scenarios arise from economic, geographic, or demographic dynamics. Each outlier event enforces a wide variety of factors which makes its analysis for healthcare policy extremely complicated. In order to extricate the process from such difficulties, in this manuscript, we present implementations of AI methods, such as Reinforcement Learning (RL), and Genetic Algorithms (GA) to evaluate different policy scenarios and point to the best possible outcome. By using the mentioned methods, we incorporate the varying factors of healthcare services along with the Traveling Salesman Problem (TSP), during conventional times as well as during outlier events to find an optimized equilibrium of resources' allocation and sharing paths. We argue that AI methods produce improved results over existing traditional methods. The research hypotheses that our study aims to evaluate are twofold: (1) AI-based methods can provide superior results (fitness value) when compared to conventional non-optimized methods for resources allocation (2) As hospitals are grappling with datasets that they collect, real-world data (collected from veteran affairs and other sources presented in section 3.1) can be used to test the presented methods and develop an interactive dashboard that presents the outcomes for decision support at a hospital.

This paper is structured as follows: the next section reviews existing resource allocation and sharing methods, as well as similar AI examples, section 3 introduces the three methods, section 4 confers the results and other discussions, and lastly, section 5 presents the conclusions.

## 2. Background and Related Work

### 2.1 Resource Allocation in Healthcare

Pandemics (such as Covid-19) and other mass casualty events place enormous demands on public health and health systems [3, 4]. They can cause a shortage of hospital beds and medical equipment. They are also likely to affect the availability of the medical workforce, since doctors and nurses likely become ill or quarantined. Such demands create the need to ration the allocation of scarce resources. If preparedness is inadequate, the foundation of all disaster response efforts can crumble with subsequent adverse patient outcomes [5].

In disasters, while some in "hot spots" have found their resources rapidly depleted, others have found themselves managing largely empty intensive care units waiting for an inevitable surge of critically ill patients. A coordinated nationwide response is needed for an effective and responsive strategy as the outlier event moves across the country. However, other than in the case of organ transplants, governments at all levels have unfortunately had little experience in engaging stakeholders in priority setting, rationing lifesaving resources, and implementing policies [6]. Since the New York attacks of 2001, the hurricane season of 2005, and the Ebola and Covid-19 outbreaks, significant resources have been used in the U.S. to improve the processes behind care provided in a disaster or pandemic [7]. However, even the well-designed allocation guidelines that have been proposed so far by various professional societies present challenging problems in real-time decision making and implementation. To help clinicians navigate these challenges, institutions have employed triage officers, physicians in roles outside direct patient care, or committees of experienced physicians and ethicists, to assist in applying guidelines and making timely decisions at the bedside [8]. A lottery system has also been proposed to better promote equity and social justice by removing the likelihood of people being given preferential treatment on the basis of social or economic advantages [9]. According to the National institutes of Health (NIH), healthcare disparity can be recognized by analyzing the relationship between healthcare utilization and household assets; outcomes of resource allocation can cause different levels of disparities [10].

One of the commonplace methods for resource allocation originated from the United Kingdom and is currently implemented in multiple health systems of Low- and Middle-Income Countries' (LMIC) such as South Africa [11]. A regional health-funding formula is created by either taxation or insurance values, this method promotes both vertical and horizontal equities: for instance, regions with the same health needs are provided with similar resources (horizontal), while regions with different health needs are provided with dissimilar resources (vertical).

Another method of resource allocation is through the use of Health Benefits Packages (HBPs). This method is most beneficial in resource-constrained settings because it offers an alternative to traditional formulas when defining area-level allocations [12]. HBPs are funded based on the expected costs of providing services and the expected target patient population.

In the UK, the Resource Allocation Working Party (RAWP) manages the process [13]. Historically, resources were allocated based on precedent, which created *geographical bias* towards areas such as London and south east England (something we address in Method #3). This produced an imbalance in the healthcare system in the UK. Recently however, health resources are disaggregated into a small number of disease categories that are based on the World Health Organization (WHO)'s international classification of disease. This approach allowed for the notion of "weighted capitation" to surface.

RAWP utilized a predefined set of variables as follows [14]: first, per capita need is calculated by disaggregating the population by age and gender. Corresponding healthcare utilization of each demographic group is approximated by using the national average hospital bed utilization (per-capita). RAWP broke down healthcare into a smaller number of broad categories of conditions and their index of approximate need of care. That is determined through the application of standardized mortality ratios and the population of geographical areas. The formula that was generated from this process relies on five variables; illness ratios, mortality ratios, proportions of economically active yet unemployed citizens, proportions of pensionable-age and "living-alone" citizens, and the proportion of dependents per household.

In the US, the challenge of healthcare expenditures has been a major part of policy debates. For example, nearly half of Americans have at least one chronic condition. Direct medical costs for chronic conditions are >$750 billion annually [15]. Out of the four possible healthcare models for a country (The Beveridge, Bismarck, Out of Pocket, and National Insurance), a model such as the Affordable Care Act (ACA – i.e. Obamacare) provides guarantees for the pursuit of preventive healthcare, but no pointers to healthcare access, resource allocation, or other geographical-relevant issues such as medical deserts in the U.S. Additionally, in recent years, the ACA has been deployed with high costs, the program has low adoption rates, and it still suffers from partial public's rejection.

Based on our extensive search for a "gold standard" in medical resources sharing, no standard was found that is agreed upon in the research community or in practice. Generally, every hospital system has their own network and resources, especially in the case of private institutions. In the case of public institutions, such as veteran affairs hospitals presented in our study, resources sharing is based on policies by the government, and is mostly an opinion/expertise-based process that is maneuvered based on the context. Even in AI research, no existing methods are found that are *dedicated* to healthcare resources management. A key objective of quantitative healthcare economic analyses is to uncover relationships – e.g. number of beds, patients, and staff – for use in making predictions or forecasts of future outcomes. However, current systems that generate forecasts for decision making tend to use ad hoc, expert-driven, personal partisanship, linear, factor and non-linear models. Multiple factors influencing healthcare economics in outlier scenarios arise from nutritional, geographic, or demographic variables and each policy comes with a wide variety of such factors. Work presented in this manuscript aims to facilitate that and provide an intelligible alternative.

### 2.2  Review of Existing AI Deployments

As it is established, measuring the *health of economy* is a long-lasting science; however, only recently, the *economy of health* is a gaining traction and is a rising research area. Successful national models in healthcare have varied between public and private. Previous studies [16] have shown that policy making for resource allocation is complex and requires thorough data analysis. AI, namely, RL and GA can provide improved suggestions over existing traditional methods. During major economic shifts, AI can capture hidden trends and provide real-time recommendations that conventional heuristic-based approaches may fail to arrive at. However, the application of AI in healthcare economics is still at its inception; in Table 1 we review "similar" methods and illustrate a chronological review (1982-2020) of RL methods applied in comparable decision-making scenarios.

**Table 1.** RL for decision making

| Year | Author(s) | Description |
| --- | --- | --- |
| 1982 | Boyan Jovanovic | Applied RL with the use of Bayesian Learning to study single firm dynamics. |

| 2001 | John Moody and Matthew Saffell | Implemented an adaptive algorithm: Recurrent RL with the utilization of Q-Learning for optimizing portfolios and asset allocations. |
|------|-------------------------------|----------------------------------------------------------------------------------------------------------------------------------|
| 2007 | Michael Schwind | Applied RL and combinatorial auctions to bidding decision problems. |
| 2014 | Neal Hughes | Author introduced methods based on Q-iteration and a batch version of Q-Learning to solve economic problems. |
| 2016 | Koichiro Ito and Mar Reguant | Use of RL to characterize strategic behavior in sequential markets under imperfect competition and restricted entry in arbitrage. |
| 2016 | Yue Deng et al. | Implemented concepts from Deep Learning (DL) and RL for real-time financial signal representations and trading. |
| 2017 | Han Cai et al. | Built a Markov Decision Process framework for learning the optimal bidding policy and optimize advertising. |
| 2017 | Saud Almahdi and Steve Y. Yang | Developing risk-based RRL portfolios for rebalancing and market condition stop-loss retraining mechanism. |
| 2018 | Jun Zhao et al. | Applied Deep RL for bidding optimization in online advertising. |
| 2018 | Thomas Spooner et al. | Authors provided one of the solutions for the market making trading problem by designing a market making agent using temporal-difference RL. |
| 2018 | Zhuoran Xiong et al. | Authors implemented Deep Deterministic Policy Gradient (DDPG) methods based on deep RL for finding the best trading strategy in complex and dynamic stock markets. |
| 2019 | Haoran Wang and Xun Yu Zhou | Authors achieved the best tradeoff between exploration and exploitation using an entropy-regularized relaxed stochastic control problem using RL. |
| 2019 | Olivier Guéant and Iuliia Manziuk | Authors proposed a new approach of implementing model-based approximations of optimal bid and ask quotes for bonds. |
| 2019 | Xinyi Li et al. | A new DDPG technique which incorporates optimistic or pessimistic deep RL for the portfolio allocation tasks. |
| 2019 | Yuming Li et al. | Deep Q-Network (DQN) for decision-making. |
| 2020 | Bastien Baldacci et al. | Authors designed approaches to approximate the financial market and other optimal controls for real-time decisions. |

As shown in Table 1, different researchers have tried applying RL methods to assisting with decision making, but to the best of our knowledge, none have provided AI-based methods for policies and decisions for resource allocation during outlier events. In this project, we present AI-based methods for that goal.

## 3. Methods

The results of data mining endeavors are majorly driven by data quality. Throughout these deployments, serious show-stopper problems are usually unresolved, such as: data collection ambiguities, data imbalance, hidden biases in data, the lack of domain information, and data incompleteness. In a traditional data science lifecycle, outliers, bias, variance, boosting, over and under sampling, and data wrangling are measures that can be tuned to mitigate output quality issues and misinformation. In our work, we performed data collection from multiple sources to mitigate issues relevant to one dataset. Additionally, normalization, descriptive analytics, hyper-parameter selection, and other quality-control methods are implemented.

### 3.1  Data Collection

Real-world data of this study are collected from Centers for Medicare and Medicaid Services (CMS) including data on patient complications and deaths, hospital general information, cost, and value of care for each hospital [17]. Additionally, data for hospital beds are collected from Open Data DC - the Office of the Chief Technology [18]. Covid-19 patients' data are collected from Centers for Disease Control and Prevention (CDC) [19]. CDC data are important in this context because hospitals in the U.S. follow CDC regulations when it comes to reporting healthcare-related practices. U.S. Veterans Affairs (VA) medical centers information is collected as well from the U.S. Department of Homeland Security [20]; the VA system constitutes the main real-world example of this paper – due to federal regulations, patients' data and hospital network information are scarce and difficult to acquire. Nonetheless, VA hospital performance star rating is also gathered [21] to be used as part of the input for training the models. Data from the U.S. Census Bureau provided the files for geo-visualization, and other studies [22] have provided the estimated recovery rates for Covid-19 patients based on gender and age group [23].

For all data sources, normalization is performed to standardize and smooth the statistical distributions of different datasets. The values are scaled to match a range. For instance, data are normalized by performing simple linear scaling from the original scale to a 0-1 scale, or a 0-100 scale. LinearScaling (an R package) is used for normalizing the range of values of a continuous variable using a linear scaling within the range of the variable, using:

```
LinearScaling(x, mx = max(x, na.rm = T), mn = min(x, na.rm = T))
```

Where *x* is a a vector with numeric values, *mx* is the maximum value of the continuous variable being normalized (defaults to the maximum of the values in *x*) and *mn* is the minimum value of the continuous variable being normalized.

All datasets, code, and scripts used for the experiment are available on a GitHub public page: https://github.com/ferasbatarseh/EconHealthAI

### 3.2  Method #1: Resources Management Using RL

RL is a subfield of AI, it involves the process of learning of an agent by using a trial and error approach that maximizes rewards by interacting with dynamic environments [24, 25]. RL is different from supervised and unsupervised learning. In supervised learning, a dataset is available with pre-defined labels that represent knowledge; while in unsupervised learning, the labels are not initially provided, and the outputs are not pre-determined. In contrast, RL is *goal-oriented*, it processes learning from the interactions as compared to other approaches of AI [26]. Chadès et al. (2017) noted the application of RL to decision making such as in the forest management problem; which inspired part of the work presented here (Method #1). We use a form of RL referred to as: Markov Decision Process (MDP), which is suitable for timeseries data and doesn't require a simulation environment like Q-Learning (another form of RL). Forest management consists of two *contrasting* objectives: maintaining an old forest for wildlife preservation, and/or making money by selling the forests' wood [27]. This tradeoff is often the case with decision making. The concept can be applied to healthcare resource allocation as follows: keep the resources (i.e. conserve wildlife), vs. share resources with other hospitals in your region (i.e. cut the trees). Such decision questions exacerbate during externalities -such as outlier events- that directly affect healthcare services and other hospital processes. In economics, an equilibrium of economic outcomes is usually modeled to study tradeoffs and support policy and decision making.

In optimization problems, algorithms are performed in an iterative fashion until the best result is found. This concept works well with RL [28]. In RL, based on the policy chosen, the agent will take an action, the action will change the environment, and the outcome is then evaluated. As the environment changes, the policy will be updated based on a reward or a punishment due to the policy change, and then the model will make a transition into a "better" new state [29].

In the RL model developed for resource allocation and sharing, multiple stages are deployed. At the beginning of each stage, an action is selected and evaluated by the algorithms (good decisions are rewarded, and "bad" decisions are punished). For instance, in the forest example:

- If the action is "cut" – which equates to the "share" option in our study, the loop goes back to stage 1; otherwise it moves to the next stage.
- If a wildfire occurs, the loop goes back to stage 1 – requires a different decision process.
- The algorithm rewards the decision when the forest trees are old, and presents a cut or wait (which equates to "idle" in section 4 – the results of our study) suggestion, which in turn, each has a different reward.
- The goal is to maximize the reward by choosing the appropriate action at each stage.

The initial values for rewards and the probability of wildfire are used through the `MDPtoolbox` package in R [30]. Additionally, a real-time web application is developed to change these values in real-time and to understand how actions and rewards changes for each state based on the "wildfire" probability. Execution data are generated using the `mdp_example_forest()` function from within `MDPtoolbox`. The tool, results, and other outcomes are presented in the results section.

### 3.3   Method #2: Resources Management Using GA

GAs is a subfield of evolutionary algorithms; they are inspired by the process of biological natural selection. By using bio-inspired methods such as crossover, mutation, and selection, GAs have been found to be one of the relatively effective algorithms for optimization problems. In this study, we deploy GAs independently, as well as in coordination with algorithms such as TSP [31] - an R package, `GA` is used. There are three types of GAs: binary, real-valued, and permutation [32, 33]. Binary GAs are used to solve binary representation of decisions such as whether the hospital should share resources or not, while permutation involves re-ordering a list of objects, which is suitable with TSP for choosing/sorting which hospitals are best to share with due to distance and other external factors – we experiment with both methods (permutation and binary).

In GA, the input data constitutes the initial random population, or as it is called: the first generation. The fitness of the generation will be measured, "parents" will be randomly selected based on a fitness function. During the reproduction state, by crossing over of "genes", two "children" will be produced, and mutations occur with a given rate. If a mutation happens, one gene will be randomly changed, which then leads to a new generation. The process continues until the termination condition is met and the population is more "fit" [34].

GAs are controlled by multiple hyperparameters, we set the following values (deemed as defaults) throughout: Probability of mutation: 50%; Population size: 1000; Maximum iterations: 250 (without improvement); Default iterations: 1000.

Healthcare variables that are inputted to the GAs as determinants of generational updates are: Hospital Performance, Death rates, and Number of beds. Additionally, two types of fitness functions are used for comparison: a fitness function with all the mentioned variables + Hospital ratings (FF1), and fitness function without hospital rating (FF2). Based on Batarseh et al. (2020), hospital ratings, reputation and ranking has a major effect on the quality of service [35]. Therefore, we aimed at exploring that aspect.

In both fitness functions: the following constants are deployed: α, β and γ. The constants aim to weight each input variable differently (if needed according to policy). Data used in this analysis have been normalized. Fitness functions 1 (FF1) and 2 (FF2) are presented below:

$$FF1 = Hospital\ Rating\ \times (\frac{\alpha \times Number\ of\ Beds}{\beta \times Death\ Rate} + \frac{1}{\gamma \times Cost}) \tag{1}$$

$$\tag{2}$$

$$FF2 = \frac{\alpha \times Number\ of\ Beds}{\beta \times Death\ Rate} + \frac{1}{\gamma \times Cost}$$

The quality of service is an important indicator of whether the hospital is ready for outlier events and whether it is well equipped. However, if Methods #1 and #2 can aid with presenting a sharing recommendation, how can a hospital manager find the best/nearest/readiest hospital to share with or share from? Method #3 addresses that question.

### 3.4   Method #3: Resources Management Using TSP

TSP is an important combinatorial optimization problem in the field of theoretical computer science and operations research [36]. Even though the purpose statement of TSP is simple: "to find the best route to visit each location once and return to the original position". TSP belongs to a family of non-deterministic polynomial-time problems (NP) [37, 38].

By using the concepts of TSP, different factors have been implanted into the traditional algorithms and used by a fitness function through GA. The goal is to optimize the best route for resources delivery during outlier events. That is while not merely considering distances, but also other factors, such as availability of resources. Due to the lack of patients' locations at the city and county levels, the algorithm is not based on specific addresses, rather, on hospital geographical locations and state centers (by longitude and latitude). We began by using the state center. The variables were used to create an index to maximize "good" sharing of resources and minimize "bad" resources' allocation. TSP is coded into two fitness function of the GA algorithm (to allow the GA to consider distance and traveling optimizations as it is parsing and learning from the data), one with healthcare variables (FF3), and other with geographical distance (FF4). Both formulas are shown below:

$$FF3 = \frac{Patient}{Cost \times Distance \times Hospital\ Performance\ Rating} \quad\quad (3)$$

$$FF4 = \frac{1}{Distance} \quad\quad (4)$$

FF3 variables have data collected from Covid-19 patients and scenarios to illustrate an outlier situation. Cost in the fitness functions is based on the following formula:

$$(5)$$
$$Cost = avg.\ recovery\ duration \times avg.\ hospital\ cost\ per\ day$$

In the experiment, only contiguous U.S. states (excluding Alaska and Hawaii) and the District of Columbia have been included. For FF4, due to the lack of patients' data in Minnesota and Wyoming, these two states are excluded as well. GAs are controlled by multiple hyperparameters, we set the following values throughout, by trial-and-error, they provided the best statistical accuracy:

   a) Probability of mutation: 20%
   b) Population size: 100
   c) Maximum number of iterations: 5000
   d) Default iterations: 1000

In addition to the `GA` package, two more R packages are used for mapping: `rgdal` is used to collect data from the Geospatial Data Abstraction Library (GDAL), and `tidygeocoder` was used for mapping the results [39, 40].

Results for FF3 and FF4 as well as all three methods (#1, #2, #3) are presented in the next section.

## 4. Models' Results and Discussions

This section presents the results of the three methods (RL, GA, and TSP) and discusses the outcomes in light of deployability and policy.

### 4.1 RL Results

Method #1 is deployed through a real-time R-Shiny web application (shown in Figure 1). But how does this apply to outlier events? As of 7th September 2020, there have been 6,280,400 patients suffering from Covid-19 in the U.S. [41]. To suitably manage the resources for patients, it can be essential to understand the pandemic's hot zones and severities. For this purpose, we used the Pandemic Severity Assessment Framework (PSAF) proposed by the CDC in 2014 [42]. PSAF is driven by two factors – clinical severity and transmissibility – which determine the overall impact of a pandemic [43]. The presented web application is to be used at a hospital for real-time decision-making support. The RL algorithm is connected to the application, and the model could be re-trained in real time when new data are available. A hospital manager can use the "hospitalization ratio", a parameter in clinical severity (aggregate of patients' situation) for determining the amount of resources needed. The mentioned RL stages are represented as scales based on transmissibility and clinical severity. The outcomes (i.e. actions) are assigned as either "Idle" or "Share" – which represents the recommendation that the RL algorithm generates.

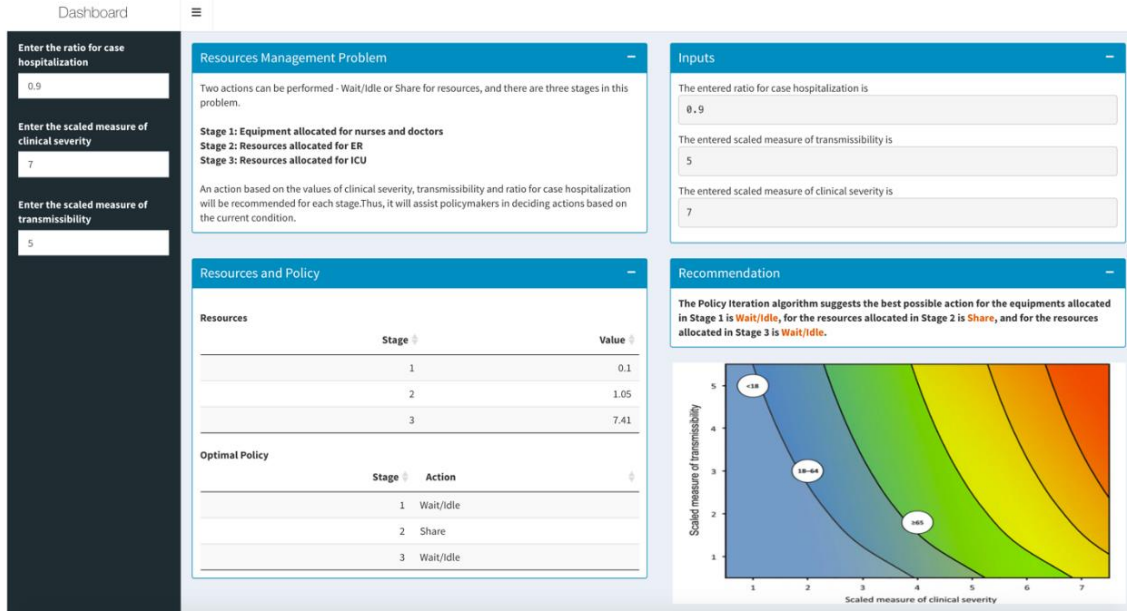

**Fig 1** Resources management dashboard using RL

The MDP RL model intakes all parameters (such as budget and staff availability) and provides a table indicating a value matrix, where every value is associated with a scale. A negative value indicates an "Ask" action for more resources, while a higher positive value indicates a surplus in resources (Table 2). RL is a dynamic classification approach, it is deemed as one of the most successful for real-time, in real environments, and for binary outcomes – we apply that for hospital resources allocation, a notion that has not been deployed prior. Our prototype can be used as an early failure detection system that can guide hospital policymakers to make resource sharing decisions in a proactive manner.

**Table 2.** Resources management decisions using RL (based on the PSAF scale)

| RL Model Inputs | RL Model Outputs |
|---|---|

| Use case | Ratio for case hospitalization | Scaled measure of clinical severity | Scaled measure of transmissibility | Stage | Value | Action | Event |
|---|---|---|---|---|---|---|---|
| 1 | 0.1 | 2 | 2 | 1 | 0.81 | Idle | *Normal* |
|   |     |   |   | 2 | 1.71 | Idle | *Normal* |
|   |     |   |   | 3 | 3.71 | Idle | *Normal* |
| 2 | 0.1 | 7 | 5 | 1 | 2.84 | Idle | *Normal* |
|   |     |   |   | 2 | 5.98 | Idle | *Normal* |
|   |     |   |   | 3 | 12.98 | Idle | *Normal* |
| 3 | 0.5 | 2 | 2 | 1 | 0.4 | Idle | *Normal* |
|   |     |   |   | 2 | 1.2 | Share | *Normal* |
|   |     |   |   | 3 | 2.8 | Idle | *Normal* |
| 4 | 0.5 | 7 | 5 | 1 | 0.88 | Idle | *Normal* |
|   |     |   |   | 2 | 2.62 | Idle | *Normal* |
|   |     |   |   | 3 | 9.62 | Idle | *Normal* |
| 5 | 0.9 | 2 | 2 | 1 | 0.1 | Idle | *Normal* |
|   |     |   |   | 2 | 1.05 | Idle | *Normal* |
|   |     |   |   | 3 | 2.15 | Idle | *Normal* |
| 6 | 0.9 | 7 | 5 | 1 | 0.1 | Idle | *Outlier* |
|   |     |   |   | 2 | 1.05 | Share | *Outlier* |
|   |     |   |   | 3 | 7.41 | Idle | *Outlier* |

## 4.2 GA Results

The algorithm in Method #2 provides policy makers suggestions of whether the hospital should request additional resources based on the current conditions of the hospital (Table 3 shows the top ten hospitals in terms of fitness). As it is noted in [35], hospital rankings are often not specific to certain goals such as ranking in readiness for resource allocation – which is part of the overall readiness for an outlier event. Using GAs (two different fitness functions) and based on the variables discussed, in Table 3, we present the readiness ranking for hospitals. If such results are embedded into a data dashboard for hospital management, managers can see which hospitals are best candidates for resources sharing during the time of need. The number of beds is generated as a decimal by the algorithm for accuracy, but could be rounded up in practice.

FF1 and FF2 use the number of beds, death rate, cost, and hospital ratings as inputs. No correlation between the input variables is observed. Table 3 presents results for Method #2. Decision 1 is based on FF1, the fitness values are shown in the FF1 column, and decision 2 is based on FF2, the values are shown in the FF2 column (both columns are highlighted). We aim to analyze the outcomes to specify which independent variables had the most effect on the outcomes, for instance, methods such as *Gain*, *DeepLift*, and *Attributions* can be applied to provide such insights.

A complete results' set for all hospitals in the state of Virginia are presented in Appendix A.

**Table 3.** Top ten hospitals in terms of resource sharing readiness using GAs

| Facility Name | Rating | # of Beds | Death Rate | Cost | Decision - 1 | FF1 | Decision - 2 | FF2 |
|---|---|---|---|---|---|---|---|---|
| CENTRA | 4 | 39.1 | 10.3 | 63.6 | 1 | 29.2 | 1 | 7.3 |
| INOVA ALEXANDRIA HOSPITAL | 5 | 42.4 | 25.5 | 64.5 | 1 | 16.9 | 1 | 3.4 |
| BON SECOURS ST MARYS HOSPITAL | 3 | 68.5 | 26.1 | 77.6 | 1 | 15.7 | 1 | 5.2 |
| MARY WASHINGTON HOSPITAL | 3 | 29.3 | 13.3 | 70.2 | 1 | 13.2 | 1 | 4.4 |
| SENTARA PRINCESS ANNE HOSPITAL | 5 | 16.3 | 15.2 | 66.5 | 1 | 11.4 | 1 | 2.3 |
| INOVA FAIR OAKS HOSPITAL | 5 | 33.7 | 30.9 | 69.2 | 1 | 11.3 | 1 | 2.3 |
| SENTARA NORFOLK GENERAL HOSPITAL | 3 | 64.1 | 34.5 | 54.8 | 0 | 11.0 | 1 | 3.7 |
| WINCHESTER MEDICAL CENTER | 4 | 51.1 | 43.6 | 50.3 | 0 | 9.4 | 0 | 2.4 |
| VIRGINIA HOSPITAL CENTER | 5 | 33.7 | 38.8 | 68.8 | 1 | 9.1 | 1 | 1.8 |
| RESTON HOSPITAL CENTER | 4 | 16.3 | 17.0 | 66.7 | 1 | 8.2 | 1 | 2.1 |
| INOVA LOUDOUN HOSPITAL | 5 | 23.9 | 30.9 | 49.7 | 1 | 8.2 | 1 | 1.6 |

In Figure 2, green bars indicate which hospitals should receive additional beds in the current context of healthcare variables (exacerbated by an outlier event), while the red bars represent the hospitals that should not receive additional beds. The height of the bars represents the fitness value of the hospital, Figure 2a shows the results of FF1, and Figure 2b shows the results after using FF2. In both Figures, the blue, orange, grey and purple marks and line represent number of beds, death rate, cost, and hospital rating for each hospital, respectively. Figure 2a and 2b indicate how different measures can produce different rankings, which is commonly a major reason to considering multiple fitness functions and data management approaches as presented. The results show that by using FF1, 25% (32% for FF2) of those with high fitness values are suggested to not receive additional beds, and 8% (12% for FF2) of hospitals with low fitness values show as they should receive additional beds. FF1 performed better overall, which aligns well with studies such as in [35] and elects the function as the most accurate resource allocation that are on a national scale – other functions are expected to perform better in more limited geographies (such as state level), a question that we aim to answer in our future work.

In Method #2, based on FF1 and FF2, the algorithms provide two different suggestions (Figure 3). Even though overall the results are highly consistent between the two fitness functions, ~9% of the hospitals have opposite suggestions on whether the hospitals should request additional beds or not during an outlier event. The bars represent the fitness values; to the left is FF1 and to the right is FF2. Green bars indicate a recommendation to requesting beds, while red recommends stopping and not requesting beds.

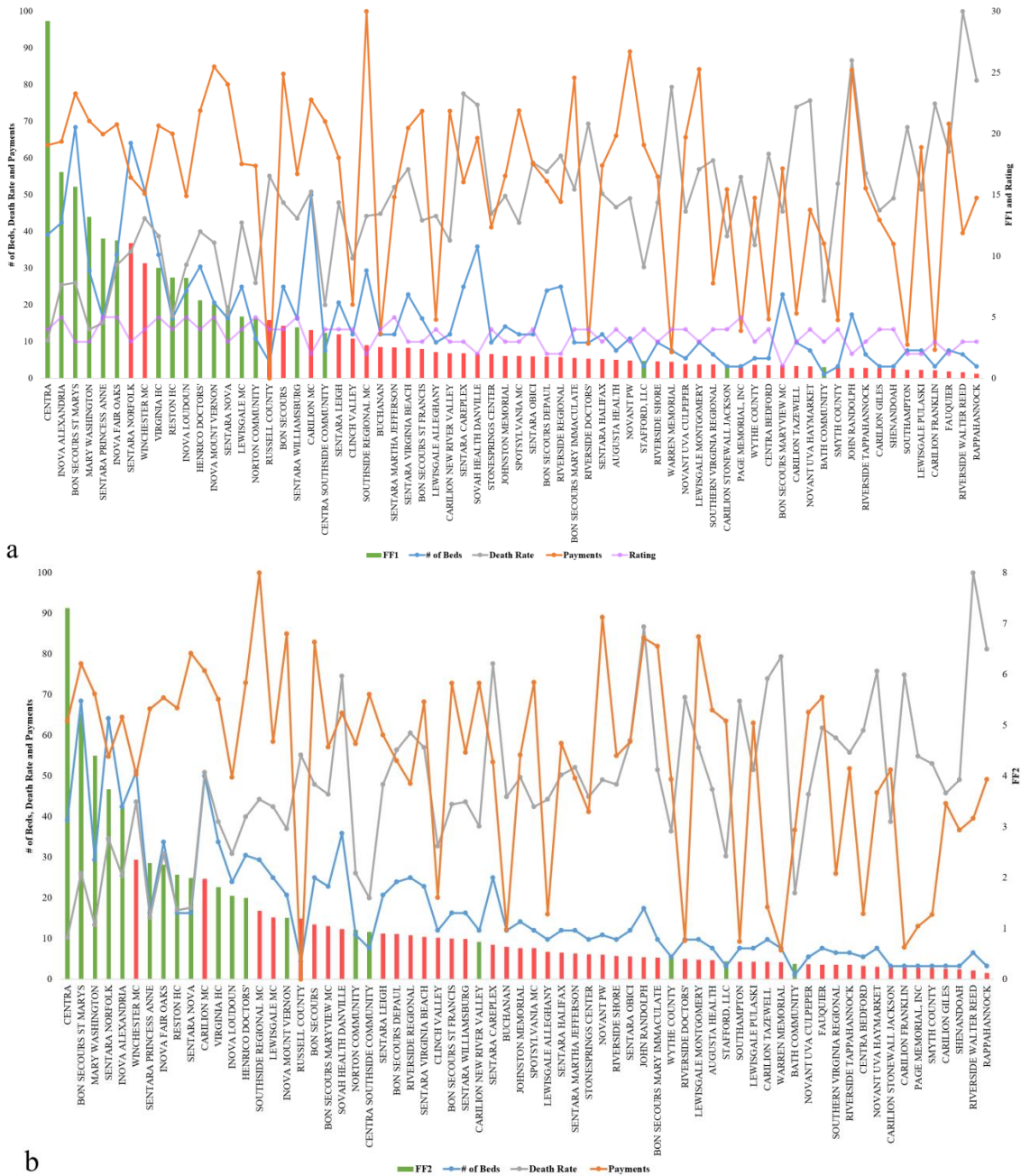

**Fig 2** Resource allocation inputs and outputs for GA using (a) FF1 and (b) FF2

In every cycle of the GA algorithm; getting new beds naturally increases the value of the input variable: number of beds; which then also increases the cost value. That leads -in the next iteration (i.e. generation) of the algorithm- to assigning some of the of the high-value hospitals a lower fitness value, or a slight increase to their fitness that is not representative in that context.

Adjusting the number of generations set in the algorithms for instance would yield slight changes to that premise.

### 4.3 TSP Results

Method #3 is especially relevant in the case of outlier events. For instance, during the current Covid-19 crises, hospitals are facing shortages on medical resources and medical professionals. A tradeoff in decisions is presented as follows: delivering the necessary limited resources to hospitals in the shortest amount of time, saving as many lives as possible, and doing so at a relatively low cost.

The fitness value of the result can vary based on different factors, such as the total number of locations, whether the data is normalized or not, and different fitness functions (presented in Table 4). It is clear that using the result with state centers (using geographical longitudes and latitudes) are more straight forward and are simpler, versus when using locations of hospitals (Figures 4a and b). However, using hospitals' locations is obviously more realistic and useful. The fitness value different is 68% higher when comparing routes between state centers (48 mainland states), or with VA hospitals (141 locations). FF3 seems to provide suggestion with lower fitness value compared to using FF4 (Table 4). But when comparing FF3 using normalized data with using unnormalized data, the fitness value improves drastically.

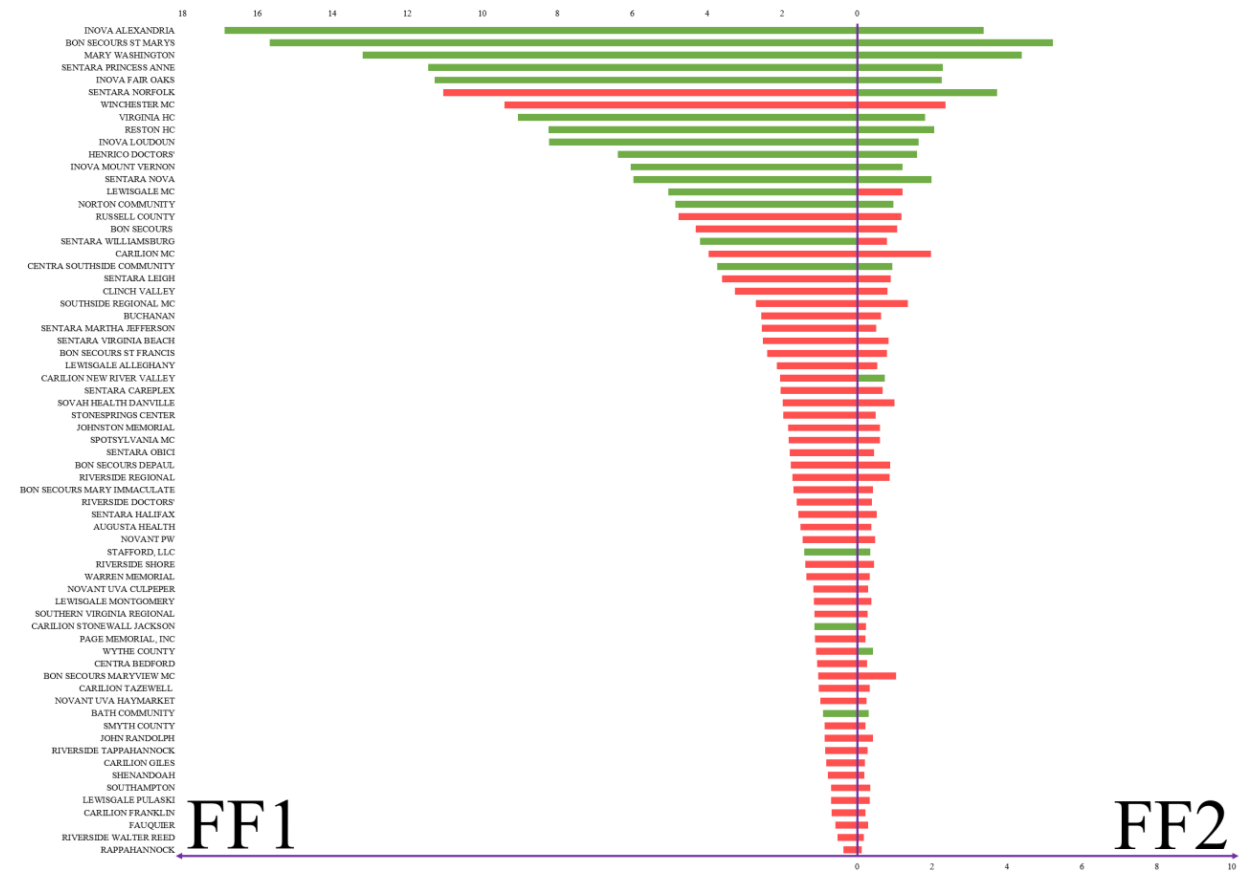

**Fig 3** Beds allocation for VA hospitals based on Method #2

**Table 4.** Fitness values with different fitness functions

| Algorithm | Fitness Value |
|---|---|
| FF4 - Medical Centers | $1.51\times10^{-3}$ |
| FF4 - State Centers | $4.74\times10^{-3}$ |
| FF3 - State Centers - Unnormalized | $1.91\times10^{-8}$ |
| FF3 - State Centers - Normalized | 1.04 |
| FF3 - State Centers – Unnormalized – K-means | $1.38\times10^{-7}$ |
| FF3 - State Centers – Normalized – K-means | 1.64 |

To improve the results further, we deployed clustering of the states by using K-means clustering (the value of *k* is selected using the elbow method, with the least sum of squared error - SSE). In reality, VA facilities are clustered into groups based on geographical locations, such as the Veterans Integrated Service Networks (VISN) system used in VA medical facilities that categorizes the U.S. to 10 geographical areas. However, by deploying K-means clustering, the number of locations/region has been changed based on statistical measures and the size of *k*, which lead to a higher average fitness value of the algorithm. Figure 4a provides route suggestion between VA medical centers using FF4 (all maps' starting point is in Pennsylvania), each point represents a VA medical centers, sized by hospital rating; Figure 4b illustrates route suggestions between state centers using FF4, each point represents the state centers. Both 7a and b use data that are not normalized; because distance is the only variable considered in FF4. In Figures 4c, d and e, state centers have been used (didn't provide results that are transferable to hospitals given the sparsity of outcomes), the size of the points indicate the sum of the hospital performance rating in the state. Figure 4c present the route suggestion using unnormalized data using FF3. Figure 4d presents the routes using normalized data with the same fitness function (FF3), and 4e presents the function with K-means clustering of the states into 4 regions (presented by 4 different colors).

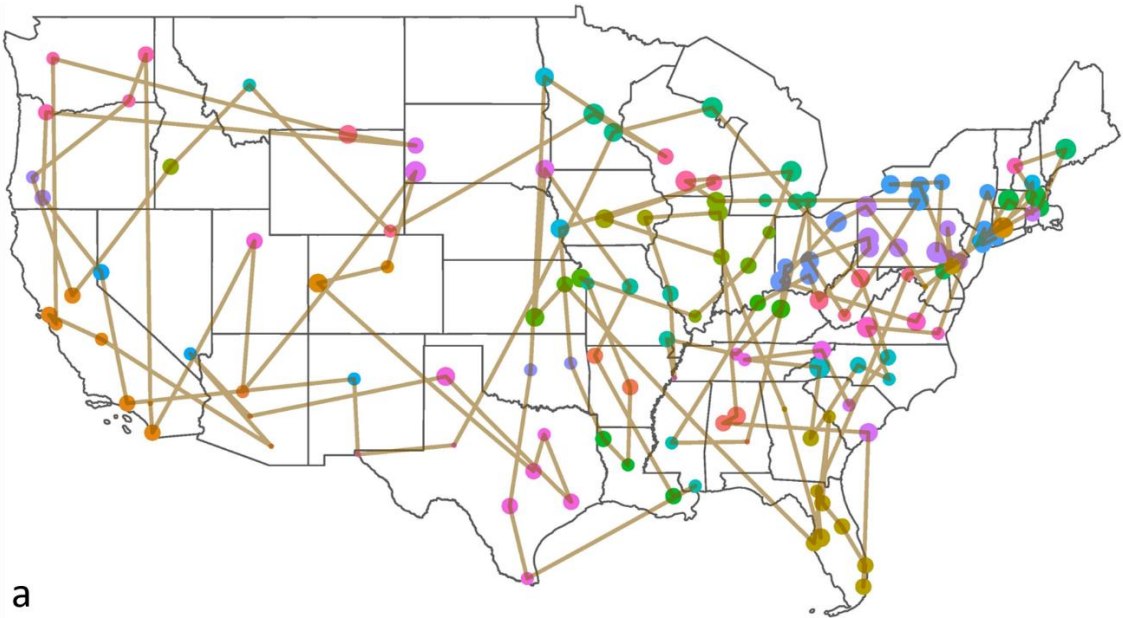

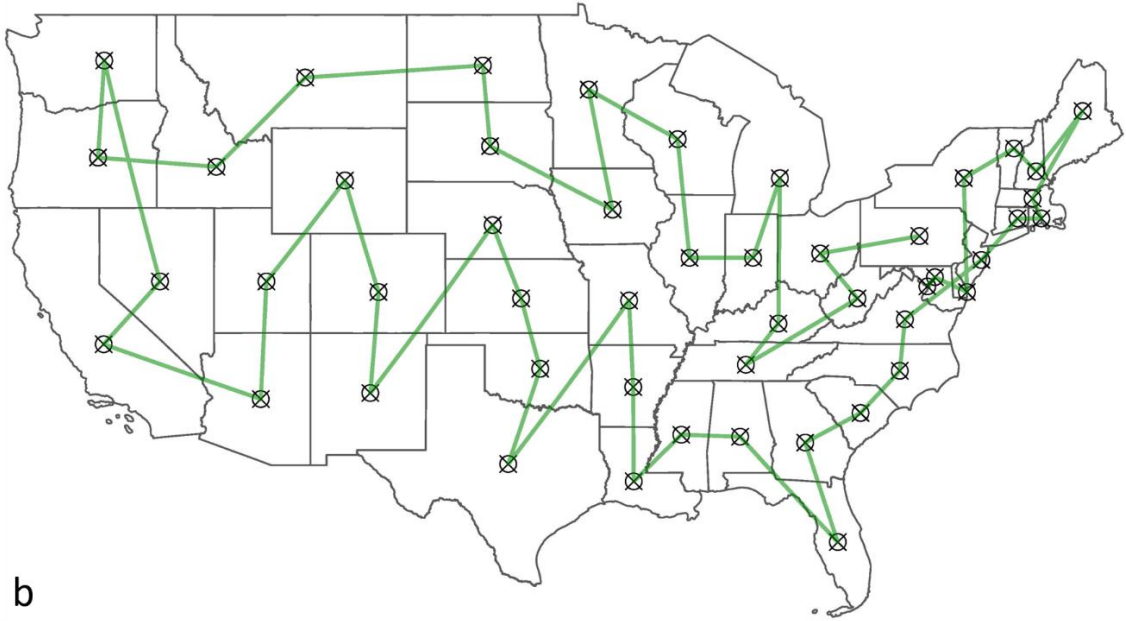

b

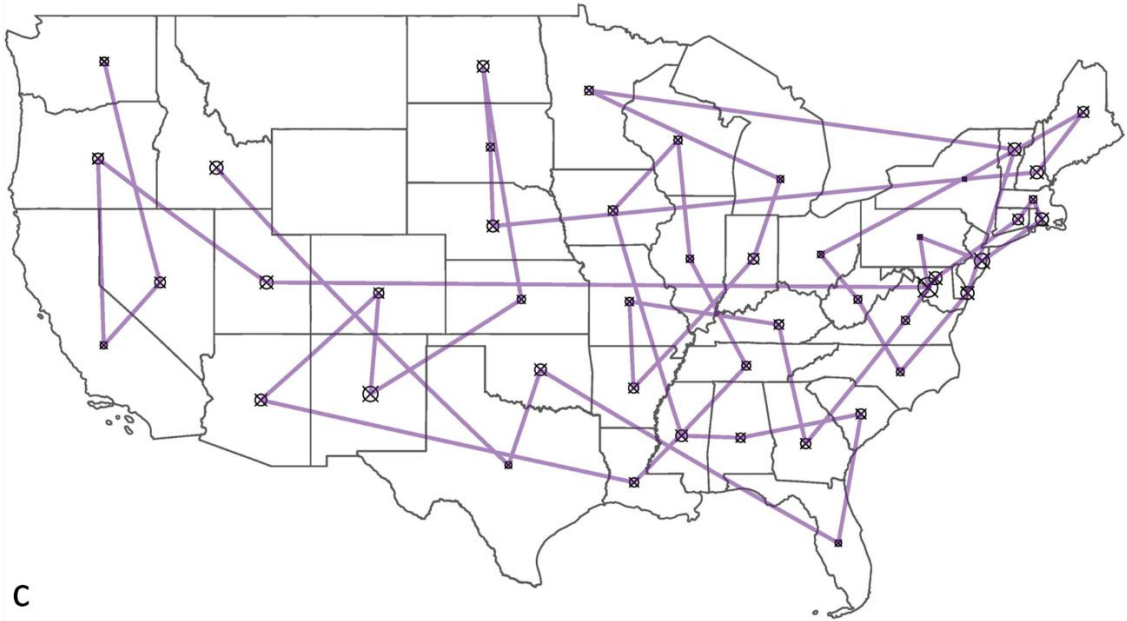

c

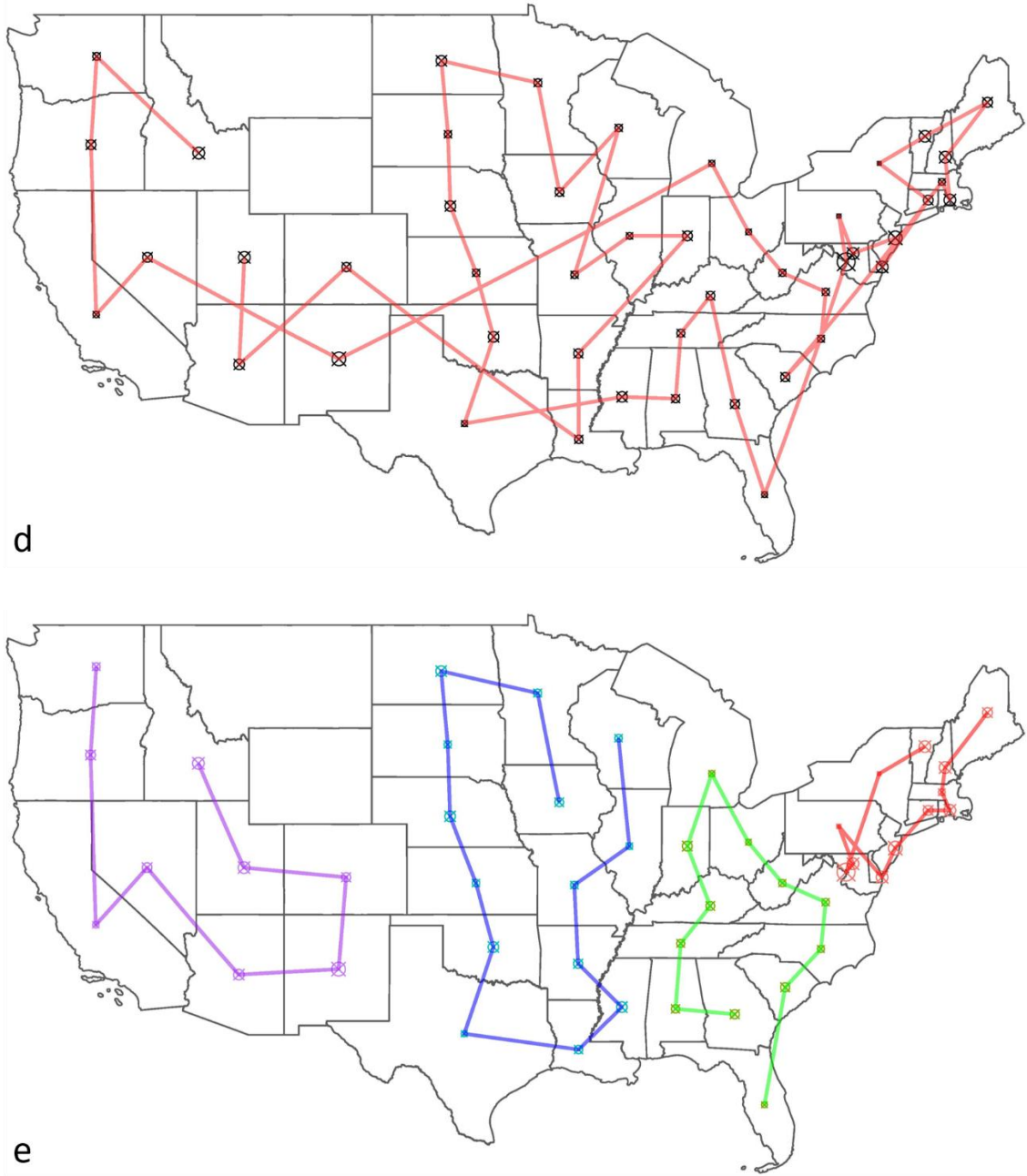

**Fig 4** Resource allocation routes using TSP with GA. (a) VA medical centers using unnormalized data with FF3; (b) state centers using unnormalized with FF3; (c) state centers using unnormalized data with FF4; (d) state centers using normalized data with FF4; and (e) state centers using k-means with FF4 with normalized data

In the resources' allocation routes experiment (Method #3), Alaska, Hawaii and other U.S. territories have been removed from the discussion due to their geographical nature. Since the distances between those locations to the contiguous U.S. are significant, the output will be greatly influenced and so we elected to remove them to avoid disturbance in results. In Figure 4, some routes show the shortest distance between two locations (a straight line). That lead to routes that go through Canada, the Great Lakes, and the Gulf of Mexico (which is troublesome). It is important to note that performing similar analysis on the state level or for within a hospital network will yield more related outcomes to hospitals; albeit a national recommendation model as the one presented in Figure 4 is beneficial for systems like the U.S. VA that are national in nature. Based on the result, using FF3 with normalized data yielded the best outcomes, and is recommended for resources' allocation routes suggestions.

## 5. Conclusions and Discussions

The resource management R dashboard using RL can help hospital administrations know whether they have additional resources that they may be able to share with others in the near future. The binary GA model provides administrators unbiased information to suggest whether the hospital should receive additional resources or not. Finally, the TSP with GA can provide unbiased route suggestion based on cost, emergency, distance and other factors. These models can be used together but may also be used independently during outlier events. Instead of using merely historical data to provide suggestions for future events, these models collect current data, and provide solutions based on the current context.

Due to the Covid-19 pandemic, telehealth service has increased dramatically (more algorithmic and real-time systems became more relevant). The algorithm for resources' allocation routes can be applied to homecare providers and others such as prescription drug delivery service. Besides distance, some patients might have more emergence conditions compared to others; therefore, using the system can provide the best route suggestion while balancing all factors.

Resource allocation strategies (even using AI) strive to maximize, treat people equally, promote and reward instrumental value, and give priority to the worst off [8]. The need to balance all these values for various interventions and in different circumstances is likely to lead to differing judgments about how much weight to give each value in particular cases (i.e. where hospital managers can then use the constants presented in our methods). The choice to set limits on access to resources is a necessary response to the overwhelming effects of mass casualty events but the question is how to do so in an ethical and systematic way, rather than basing decisions on individual institutions' approaches or a clinician's intuition in the heat of the moment [8]. For humans, that notion is difficult to measure, for an algorithm however, AI assurance methods (such as for measuring bias) could be applied to provide insights into the validity of the outcomes, a notion that we aim to experiment as part of future work.

Pandemics have shaken healthcare systems around the globe, turning attention and resources away from patients who need other types of care. Providers defer elective and preventive visits. Patients decline emergency room (ER) visits, hospitalizations, and elective procedures. Children go unvaccinated; blood pressure is left uncontrolled; cancer survivors miss their checkups. A decline of 42% was reported by the CDC in emergency room visits because of the Covid-19 pandemic [44], a decline of 42% in VA hospitals admissions [45] and 33.7% in hospital admissions for eight other acute care hospitals [46], a decline of 38% to 40% in myocardial infarctions [47], and a decline in the use of stroke imaging by 39% [48]. Sicker admissions of non-Covid-19 patients to the ICU are noted [49], confirming a trend where patients most often reached sicker health status, seeking care later, and avoiding hospitals due to fear of the Covid-19 infection. To mitigate the risks for decompensation, morbidity, and loss to follow-up that could result from delayed care for patients, providers are now converting, when possible, in-person visits to telemedicine visits [50, 51]. Our models could be merged with some telehealth services and provide a more comprehensive view to hospital management. Telehealth and other digital services come along with additional benefits for both the patient and the provider including potential faster care, cost savings, time savings, increased access to healthcare for rural areas, limiting the spread of Covid-19, and help preserve the limited supply of Personal Protective Equipment (PPE) [52]. Patient satisfaction for telemedicine care has been consistently high [53].

A recent report, evaluating 400 patients scheduled for office consultation, showed that about 95% of the patients were at increased risk for a severe outcome of Covid-19 and that 85% of them would have favored telemedical consultation during the pandemic [54]. However, there are also some potential negatives about telemedicine services, including less access for those without basic technology or computer skills, the potential of paying out-of-pocket, interstate licensure challenges and other regulatory issues that may vary by state, the need to address sensitive topics, especially if there is patient discomfort or concern for privacy, and that not all conditions are appropriate for online treatment [55, 56]. Properly trained, a learning algorithm can augment physician judgment about when to offer or subtract life-saving care and how to allocate scarce resources in challenging mass critical illness [57, 58]. We anticipate that our methods will help in these scenarios, however, the following limitations are also areas where we wish to improve on: (1) testing the system with data from other health institutions and (2) trying different AI models to evaluate if they overperform RL and GA. (3) Although healthcare data are scarce and difficult to gather, we aim to inject more health variables into our algorithms to improve their accuracy and usability. Additionally, as part of future work, (4) we will test the mentioned methods presented in real-world scenarios, at hospitals such as Massachusetts General Hospital (MGH), the U.S. VA network, and other potential locations. Besides testing feasibility, deploying the system at a hospital will help evaluate its sanity and confirm its accuracy and relevance. It is evident that human judgment will remain important and ultimate [59, 60], but it can be supported with the dispassionate independent score-keeping capabilities of AI. This will be a logical extension of the risk-assessment tools used today in medicine.

## Disclosure Statements:

**Data Availability Statement:**
The data that support the findings of this study, code, and the data dashboard are openly available in a GitHub public repository, https://github.com/ferasbatarseh/EconHealthAI

**Competing interest Statement:**
No competing interests are declared.

**Funding Statement:**
No funding was provided for this study.

**Acknowledgements:**
Not applicable

**Authors contributions:**
C.H. led the project's experiments and wrote the manuscript, C.H. also developed the maps. F.B. designed the research plan, developed the idea and its implementation. F.B. also validated and evaluated the experimental work. A.B. developed the medical analysis and helped with writing as well as provided medical and healthcare insights to the paper. A.K. and P.S. collected data, developed the AI models (RL and GA), developed the dashboard, and designed the figures. J.H. wrote several sections of the paper and developed R code for the AI models.

# References

1.  Niven PR (2002) Maximizing Performance and Maintaining Results, 2nd ed. John Wiley & Sons, Inc.

2.  McGuire PK, Silbersweig DA, Wright I, et al (1995) Abnormal monitoring of inner speech: a physiological basis for auditory hallucinations. Lancet 346:596–600. https://doi.org/10.1016/s0140-6736(95)91435-8

3.  Bell SA, Abir M, Choi H, et al (2018) All-Cause Hospital Admissions Among Older Adults After a Natural Disaster. Ann Emerg Med 71:746-754.e2. https://doi.org/10.1016/j.annemergmed.2017.06.042

4.  Christian MD, Sprung CL, King MA, et al (2014) Triage: care of the critically ill and injured during pandemics and disasters: CHEST consensus statement. Chest 146:e61S-74S. https://doi.org/10.1378/chest.14-0736

5.  Sandrock CE (2014) Care of the critically ill and injured during pandemics and disasters: groundbreaking results from the Task Force on Mass Critical Care. Chest 146:881–883. https://doi.org/10.1378/chest.14-1900

6.  Ramachandran P, Swamy L, Kaul V, Agrawal A (2020) A National Strategy for Ventilator and ICU Resource Allocation During the Coronavirus Disease 2019 Pandemic. Chest 158:887–889. https://doi.org/10.1016/j.chest.2020.04.050

7.  Romney D, Fox H, Carlson S, et al (2020) Allocation of Scarce Resources in a Pandemic: A Systematic Review of US State Crisis Standards of Care Documents. Disaster Med Public Health Prep 1–7. https://doi.org/10.1017/dmp.2020.101

8.  Emanuel EJ, Persad G, Upshur R, et al (2020) Fair Allocation of Scarce Medical Resources in the Time of Covid-19. N Engl J Med 382:2049–2055. https://doi.org/10.1056/NEJMsb2005114

9.  Silva DS (2020) Ventilators by Lottery: The Least Unjust Form of Allocation in the Coronavirus Disease 2019 Pandemic. Chest 158:890–891. https://doi.org/10.1016/j.chest.2020.04.049

10. Gómez EJ, Jungmann S, Lima AS (2018) Resource allocations and disparities in the Brazilian health care system: insights from organ transplantation services. BMC Health Services Research; London 18:. http://dx.doi.org.mutex.gmu.edu/10.1186/s12913-018-2851-1

11. Love-Koh J, Griffin S, Kataika E, et al (2020) Methods to promote equity in health resource allocation in low- and middle-income countries: an overview. Global Health 16:6. https://doi.org/10.1186/s12992-019-0537-z

12. Schreyögg J, Stargardt T, Velasco-Garrido M, Busse R (2005) Defining the "Health Benefit Basket" in nine European countries. Eur J Health Econ 6:2–10. https://doi.org/10.1007/s10198-005-0312-3

13.  Smith PC (2008) Resource allocation and purchasing in the health sector: the English experience. Bull World Health Organ 86:884–888. https://doi.org/10.2471/blt.07.049528

14.  Gillie A, Slater M, Biswas T, et al (1988) Resource Allocation in the Health Service. A Review of the Methods of the Resource Allocation Working Party (RAWP). Economic Journal 98:224–265. https://doi.org/10.2307/2233541

15.  Batarseh FA, Ghassib I, Chong D (Sondor), Su P-H (2020) Preventive healthcare policies in the US: solutions for disease management using Big Data Analytics. J Big Data 7:38. https://doi.org/10.1186/s40537-020-00315-8

16.  Friedman EJ, Oren SS (1995) The complexity of resource allocation and price mechanisms under bounded rationality. Economic Theory 6:225. https://doi.org/10.1007/BF01212489

17.  Centers for Medicare and Medicaid Services (2020) 2020 Annual Hospital Compare Data

18.  Open Data DC (2020) Definitive Healthcare: USA Hospital Beds

19.  National Center for Health Statistics (2020) Provisional COVID-19 Death Counts by Sex, Age, and State | Data | Centers for Disease Control and Prevention

20.  U.S. Department of Homeland Security (2020) Veterans Health Administration Medical Facilities

21.  U.S. Department of Veterans Affairs (2018) End of Year Hospital Star Rating (FY2018) - Quality of Care

22.  Voinsky I, Baristaite G, Gurwitz D (2020) Effects of age and sex on recovery from COVID-19: Analysis of 5769 Israeli patients. J Infect 81:e102–e103. https://doi.org/10.1016/j.jinf.2020.05.026

23.  U.S. Census Bureau (2019) TIGER/Line® Shapefiles

24.  Kaelbling LP, Littman ML, Moore AW (1996) Reinforcement Learning: A Survey. arXiv:cs/9605103

25.  Szepesvári C (2010) Algorithms for Reinforcement Learning. Synthesis Lectures on Artificial Intelligence and Machine Learning 4:1–103. https://doi.org/10.2200/S00268ED1V01Y201005AIM009

26.  Sutton RS, Barto AG (2018) Reinforcement Learning: An Introduction, 2nd ed. MIT Press

27.  Cros M-J, Peyrard N, Sabbadin R (2016) GMDPtoolbox: a Matlab library for solving Graph-based Markov Decision Processes. JFRB 17

28.  Li K, Malik J (2016) Learning to Optimize. arXiv:160601885 [cs, math, stat]

29.  Mousavi A, Li L (2020) Off-Policy Estimation for Infinite-Horizon Reinforcement Learning. In: Google AI Blog. http://ai.googleblog.com/2020/04/off-policy-estimation-for-infinite.html. Accessed 30 Sep 2020

30. Chades I, Chapron G, Cros M-J, et al (2017) MDPtoolbox: Markov Decision Processes Toolbox

31. Kim E, Lee S, Kim JH, et al (2013) Implementation of novel model based on Genetic Algorithm and TSP for path prediction of pandemic. In: 2013 International Conference on Computing, Management and Telecommunications (ComManTel). IEEE, Ho Chi Minh City, pp 392–396

32. Scrucca L (2013) GA: A Package for Genetic Algorithms in R. J Stat Soft 53:. https://doi.org/10.18637/jss.v053.i04

33. Scrucca L (2017) On Some Extensions to GA Package: Hybrid Optimisation, Parallelisation and Islands Evolution. The R journal 9:187–206

34. Woodward RI, Kelleher EJR (2016) Towards 'smart lasers': self-optimisation of an ultrafast pulse source using a genetic algorithm. Scientific Reports 6:37616. https://doi.org/10.1038/srep37616

35. Batarseh FA, Yang R (2020) Data democracy: at the nexus of artificial intelligence, software development, and knowledge engineering, 1st ed. Academic Press is an Imprint of Elsevier, ISBN: 9780128189399

36. Lu J, Fang N, Shao D, Liu C (2007) An Improved Immune-Genetic Algorithm for the Traveling Salesman Problem. In: Third International Conference on Natural Computation (ICNC 2007). IEEE, Haikou, China, pp 297–301

37. Calude CS (2013) Inductive Complexity of the P Versus NP Problem. 16

38. Hillar CJ, Lim L-H (2013) Most Tensor Problems Are NP-Hard. Journal of the ACM 60:39

39. Bivand R, Keitt T, Rowlingson B, et al (2020) rgdal: Bindings for the "Geospatial" Data Abstraction Library

40. Cambon J (2020) tidygeocoder: Geocoding Made Easy

41. Johns Hopkins University & Medicine (2020) COVID-19 in the USA. In: Johns Hopkins Coronavirus Resource Center. https://coronavirus.jhu.edu/. Accessed 18 Sep 2020

42. Centers for Disease Control and Prevention (2016) Pandemic Severity Assessment Framework (PSAF). In: Pandemic Influenza (Flu) - CDC. https://www.cdc.gov/flu/pandemic-resources/national-strategy/severity-assessment-framework.html. Accessed 18 Sep 2020

43. Reed C, Biggerstaff M, Finelli L, et al (2013) Novel Framework for Assessing Epidemiologic Effects of Influenza Epidemics and Pandemics. Emerg Infect Dis 19:85–91. https://doi.org/10.3201/eid1901.120124

44. Hartnett KP, Kite-Powell A, DeVies J, et al (2020) Impact of the COVID-19 Pandemic on Emergency Department Visits - United States, January 1, 2019-May 30, 2020. MMWR Morb Mortal Wkly Rep 69:699–704. https://doi.org/10.15585/mmwr.mm6923e1

45.  Baum A, Schwartz MD (2020) Admissions to Veterans Affairs Hospitals for Emergency Conditions During the COVID-19 Pandemic. JAMA 324:96–99. https://doi.org/10.1001/jama.2020.9972

46.  Oseran AS, Nash D, Kim C, et al (2020) Changes in hospital admissions for urgent conditions during COVID-19 pandemic. Am J Manag Care 26:327–328. https://doi.org/10.37765/ajmc.2020.43837

47.  Garcia S, Albaghdadi MS, Meraj PM, et al (2020) Reduction in ST-Segment Elevation Cardiac Catheterization Laboratory Activations in the United States During COVID-19 Pandemic. J Am Coll Cardiol 75:2871–2872. https://doi.org/10.1016/j.jacc.2020.04.011

48.  Kansagra AP, Goyal MS, Hamilton S, Albers GW (2020) Collateral Effect of Covid-19 on Stroke Evaluation in the United States. N Engl J Med 383:400–401. https://doi.org/10.1056/NEJMc2014816

49.  Fadel FA, Al-Jaghbeer M, Kumar S, et al (2020) The impact of the state of Ohio stay-at-home order on non-COVID-19 intensive care unit admissions and outcomes. Anaesthesiology Intensive Therapy 52:249–252. https://doi.org/10.5114/ait.2020.98393

50.  Gadzinski AJ, Gore JL, Ellimoottil C, et al (2020) Implementing Telemedicine in Response to the COVID-19 Pandemic. J Urol 204:14–16. https://doi.org/10.1097/JU.0000000000001033

51.  Mann DM, Chen J, Chunara R, et al (2020) COVID-19 transforms health care through telemedicine: Evidence from the field. J Am Med Inform Assoc 27:1132–1135. https://doi.org/10.1093/jamia/ocaa072

52.  Kichloo A, Albosta M, Dettloff K, et al (2020) Telemedicine, the current COVID-19 pandemic and the future: a narrative review and perspectives moving forward in the USA. Fam Med Community Health 8:. https://doi.org/10.1136/fmch-2020-000530

53.  Kruse CS, Krowski N, Rodriguez B, et al (2017) Telehealth and patient satisfaction: a systematic review and narrative analysis. BMJ Open 7:e016242. https://doi.org/10.1136/bmjopen-2017-016242

54.  Novara G, Checcucci E, Crestani A, et al (2020) Telehealth in Urology: A Systematic Review of the Literature. How Much Can Telemedicine Be Useful During and After the COVID-19 Pandemic? Eur Urol. https://doi.org/10.1016/j.eururo.2020.06.025

55.  Hjelm NM (2005) Benefits and drawbacks of telemedicine. J Telemed Telecare 11:60–70. https://doi.org/10.1258/1357633053499886

56.  Ruiz Morilla MD, Sans M, Casasa A, Giménez N (2017) Implementing technology in healthcare: insights from physicians. BMC Med Inform Decis Mak 17:92. https://doi.org/10.1186/s12911-017-0489-2

57.  Alimadadi A, Aryal S, Manandhar I, et al (2020) Artificial intelligence and machine learning to fight COVID-19. Physiological Genomics 52:200–202. https://doi.org/10.1152/physiolgenomics.00029.2020

58.   Allam Z, Jones DS (2020) On the Coronavirus (COVID-19) Outbreak and the Smart City Network: Universal Data Sharing Standards Coupled with Artificial Intelligence (AI) to Benefit Urban Health Monitoring and Management. Healthcare (Basel) 8:. https://doi.org/10.3390/healthcare8010046

59.   Garratt KN, Schneider MA (2019) Thinking Machines and Risk Assessment: On the Path to Precision Medicine. J Am Heart Assoc 8:e011969. https://doi.org/10.1161/JAHA.119.011969

60.   Shah P, Kendall F, Khozin S, et al (2019) Artificial intelligence and machine learning in clinical development: a translational perspective. NPJ Digit Med 2:69. https://doi.org/10.1038/s41746-019-0148-3